\documentclass[runningheads]{llncs}
\usepackage{graphicx}
\begin{document}
\title{Vehicle Detection and Tracking From Surveillance Cameras in Urban Scenes}
%

\author{Oumayma Messoussi\inst{1}
\and
Felipe Gohring de Magalhães\inst{1} \and
Francois Lamarre\inst{2} \and Francis Perreault\inst{2} \and Ibrahima Sogoba\inst{2} \and Guillaume-Alexandre Bilodeau\inst{1}
\and Gabriela Nicolescu\inst{1}
}
\authorrunning{O. Messoussi et al.}

\institute{
Polytechnique Montreal, Montreal, Quebec H3T 1J4, Canada\\
\email{\{oumayma.messoussi,felipe.gohring-de-magalhaes,\\guillaume-alexandre.bilodeau,gabriela.nicolescu\}@polymtl.ca} 
\and
Cysca Technologies, Montreal, Quebec H2Y 1N9, Canada\\
\email{\{francois.lamarre,francis.perreault,ibrahima.sogoba\}@cysca.ca} 
\url{https://www.cysca.ca/} 
}
\maketitle              
\begin{abstract}
Detecting and tracking vehicles in urban scenes is a crucial step in many traffic-related applications as it helps to improve road user safety among other benefits. Various challenges remain unresolved in multi-object tracking (MOT) including target information description, long-term occlusions and fast motion. We propose a multi-vehicle detection and tracking system following the tracking-by-detection paradigm that tackles the previously mentioned challenges. Our MOT method extends an Intersection-over-Union (IOU)-based tracker with vehicle re-identification features.  This allows us to utilize appearance information to better match objects after long occlusion phases and/or when object location is significantly shifted due to fast motion. We outperform our baseline MOT method on the UA-DETRAC benchmark while maintaining a total processing speed suitable for online use cases.

\keywords{Multi-object tracking  \and Tracking-by-detection \and Object re-identification.}
\end{abstract}

\section{Introduction}

In smart cities, intelligent transportation systems that rely on cameras and other sensors play a critical role in ensuring road users safety. Autonomous vehicles, for example, require superior computer vision algorithms to comprehend their surroundings and to have the ability to interact with them. In order to navigate between two points, they must be able to recognize road signs, traffic signals, people, and automobiles. Detecting traffic violations and managing parking spaces are illustrations of other applications. Automated parking management, for example, necessitates recognizing when vehicles enter the parking area and following their path until they reach the parking place. This aids in saving expenses and eliminating manual work. The ability to describe a vehicle trajectory is also important for the user's safety because it tracks their movement and, as a result, can detect how it interacts with surrounding moving objects in order to avoid crashes. Therefore, video monitoring has several advantages for both road users and those in charge of them.

When tracking vehicles, a surveillance system can offer information about an object category, location, and trajectory utilizing multi-target detection and tracking algorithms. We aim to develop a method to detect and track multiple vehicles 
in single-camera views in an online use case. We also tackle challenges like motion-related displacements and long-term occlusions.

To solve this, there are numerous approaches to this problem, which we review in section \ref{relwork}. We approach this problem from a tracking-by-detection perspective, in which detection and tracking are performed in two steps. The first task, known as multi-object detection, identifies and locates vehicles in a video frame. Multi-object tracking is the next phase, which assigns unique identifiers to detected objects, extracts information describing them, and connects detections in each frame to create trajectories.

Within this approach, some methods use the IOU metric to match detections through time by measuring the overlap between candidate bounding boxes \cite{1517Bochinski2017,1547Bochinski2018,Bewley_2016}. The main appeal of this approach is its low execution time. These methods can process a high number of frames per second making them well suited for real-time or online use cases. Yet there is still room for improvement by incorporating other types of features to describe the candidate objects. Therefore, in our work, we build upon V-IOU \cite{1547Bochinski2018} and add an appearance feature component from a vehicle re-identification model \cite{reid} thus yielding more robust results in challenging scenes with large displacements due to fast motion and long-term occlusions. 

Our contributions in this work are the following:
\begin{itemize}
    \item Performance improvement in terms of multi-object tracking metrics by extending data association with re-identification features.
    \item Online/fast end-to-end detection and tracking achieved through the choice of models as well as the design of the data association step. 
\end{itemize}

\section{Background and Related Works}
\label{relwork}

\subsection*{Challenges in MOT}

Multi-object tracking is an active topic in computer vision research, owing to the numerous challenges it presents. A tracking model faces difficulties due to potential crossing tracks, which can alter identities or produce gaps in tracks, similar to how a human struggles to keep track of multiple things in a video sequence. The following are some of the most significant issues encountered:

\begin{itemize}
    \item \textbf{Occlusion}: Another target or a background item can partially or completely obscure a target's view. During the occlusion phase, the object features become erroneous or outdated. The features of the front target interfere with those of the occluded target when partial occlusion is used. Full occlusion, on the other hand, renders the features of the target irreversible.
    \item \textbf{Fast motion}: When an object moves at a high speed in a video, its position changes notably between frames. As a result, spatial data about a target becomes unreliable. Fast movement can also lead to motion blur, causing appearance descriptors to be inaccurate.
    \item \textbf{Scale changes}: When a target travels closer or further away from the camera, its scale changes, and its size changes as well. As a result, the tracking technique must react to these changes and update the target bounding box in order to prevent losing information about the whole object or accidentally adding background data.
    \item \textbf{Illumination variations}: Changes in illumination surrounding the targets, such as those caused by the weather, might affect the look of the objects. 
    A hazy sky or a night setting may also drastically alter the appearance of objects. In the dark, for example, a vehicle headlights seem like spotlights on the camera, preventing it from properly capturing the vehicle. These flaws make tracking more difficult and make identity switches more frequent.
    \item \textbf{Target similarity}: Because of the significant resemblance in appearance, tracking several items of the same category increases the complexity of the task. This makes distinguishing between two or more targets and maintaining their individual identities difficult.
\end{itemize}

Object detection algorithms have improved dramatically in recent years owing to advances in deep learning and sophisticated CNN architectures \cite{ren2016faster,he2018mask,redmon2016look,zhou2019objects,perreault2020spotnet}. To take advantage of the excellent detection results, the majority of top-performing MOT techniques employ the tracking-by-detection approach. In this method, an object detector is applied to each frame, and the MOT method is then used to find the best match between objects.

As opposed to batch tracking methods like IHTLS \cite{Dicle_2013_ICCV}, revisited JPDA \cite{Rezatofighi_2015_ICCV} and GOG \cite{GOG}, which process a number of frames or a whole video sequence before giving results, online MOT techniques process one frame at a time without having visibility or knowledge about future frames. As a result, online MOT techniques are more naturally suited to real-time systems, yet not all online MOT algorithms are real-time. Due to access to information from previous and future frames, offline MOT techniques naturally perform better in terms of MOT performance measures. Offline techniques can use this information to link/fill gaps in fragmented tracks, for example. Consequently, online MOT is a more difficult task by definition.

\subsection*{Optimization approaches}

Detection-based tracking method frequently present the following four components: object detection, object state propagation, data association, and track lifespan management. SORT \cite{Bewley_2016}, an early online tracking method, proposed a simple approach for enhancing frame-to-frame association. They accomplish so by employing the Faster R-CNN object detection model, then modeling the target state by spatial and motion information. The IOU metric is used to associate detections to tracks by measuring the amount of overlap between bounding boxes. When a match is made, the target state is updated using the detection information and a Kalman \cite{kalman} motion model. The Hungarian algorithm \cite{hungarian} is applied to solve the cost matrix computed with the IOU scores, with a minimum overlap threshold to select assignments. While this approach has been reported to analyze video sequences at a speed of up to 260 frames per second (FPS), it struggles with long-term occlusions, resulting in an increase in the number of ID change errors. DeepSORT \cite{wojke2017simple} was later proposed as an improvement by adding appearance information extracted from a person re-identification (ReID) model. This combination of spatial cues from the IOU score and the ReID features, give results in terms of tracking metrics that are more robust.

Similarly to SORT, the data association component of the IOU tracker \cite{1517Bochinski2017} is reduced to a simple IOU between detected bounding boxes in frame pairs. The authors argue that having strong high-accuracy detections enables for simpler tracking techniques to reach greater processing rates while still competing with more complex MOT approaches in terms of tracking performance, without relying on visual or motion characteristics. Yet, a small number of missed detections can dramatically increase the number of ID switches and fragmentation errors. Hence, the IOU tracker was improved later by adding visual object tracking in V-IOU \cite{1547Bochinski2018}. Forward visual tracking predicts a target location when detections are not matched, for a fixed number frames $ TTL $ (time to live). When new tracks are created, backward visual tracking is used to see if they can be linked to previously terminated tracks. V-IOU tracker achieves a good trade-off between speed and tracking performance with inference speeds of over 200 FPS, thus maintaining its primary selling feature. However, this technique does not include appearance elements for full occlusion circumstances when the visual tracker fails, therefore there is still space for development. The IOU-based association cost is also greatly affected by displacements caused by fast motion. When an item travels quickly between two frames, the bounding boxes may be widely apart, resulting in little or no overlap. This lowers the IOU score which in turn raises the association cost, making it impossible to link the boxes even if they represent the same target.

\subsection*{Neural network approaches}

While previous approaches represented data association as an optimization problem, works like DAN \cite{DAN} and DMAN \cite{zhu2019online} use end-to-end neural networks to tackle the challenge. DAN (Deep Affinity Network) learns the appearance cues of objects and their affinities at different abstraction levels and then computes correlations between pairs of features vectors. DMAN (Dual Matching Attention Networks) proposes the use of a cost-sensitive loss function, as well as temporal and spatial attentions, in a single object tracker for each candidate detection in a frame. While these methods achieve comparable performances with top performers on some MOT benchmarks, they significantly compromise on speed, with around 6 FPS for DAN and 0.3 FPS for DMAN.

Recurrent neural networks are also used in other works \cite{KieritzHA18,10.5555/3298023.3298181} to describe temporal relationships between targets. Recurrent networks are used to combine detections into tracks based on different features (spatial, appearance or motion cues). The initial findings in \cite{KieritzHA18} reveal that this technique has a significant rate of false positives and ID switches, thus the authors recommend using ReID cues for occluded objects. 

Braso et al. \cite{braso2020learning} suggested a graph-based method that makes use of the MOT usual network flow. Instead of estimating pairwise costs, the neural solver for the MOT approach learns to predict final trajectories directly from graphs. To accomplish so, a message passing network (MPN) that employs appearance and geometry characteristics is used to train on the MOT graph domain. 
Lee et al. suggested FPSN \cite{8587153}, or Feature Pyramid Siamese Network, that employed a Feature Pyramid Network to extract target features from multiple levels, resulting in a multi-level discriminative feature. FPSN-MOT adds spatio-temporal motion characteristics to address the absence of motion information in traditional Siamese networks. 
Osep et al. presented Track, then Decide: Category-Agnostic Vision-based Multi-Object Tracking \cite{osep2017track}, a tracking by segmentation approach based on pixel-level information. To create object segmentations, this technique is model-free and employs category-agnostic picture segmentation. Objects are then tracked by using semantic information to conduct mask-based association on pixels.

\section{Proposed Solution}

\begin{figure}[h]

\centering
\includegraphics[width=\linewidth]{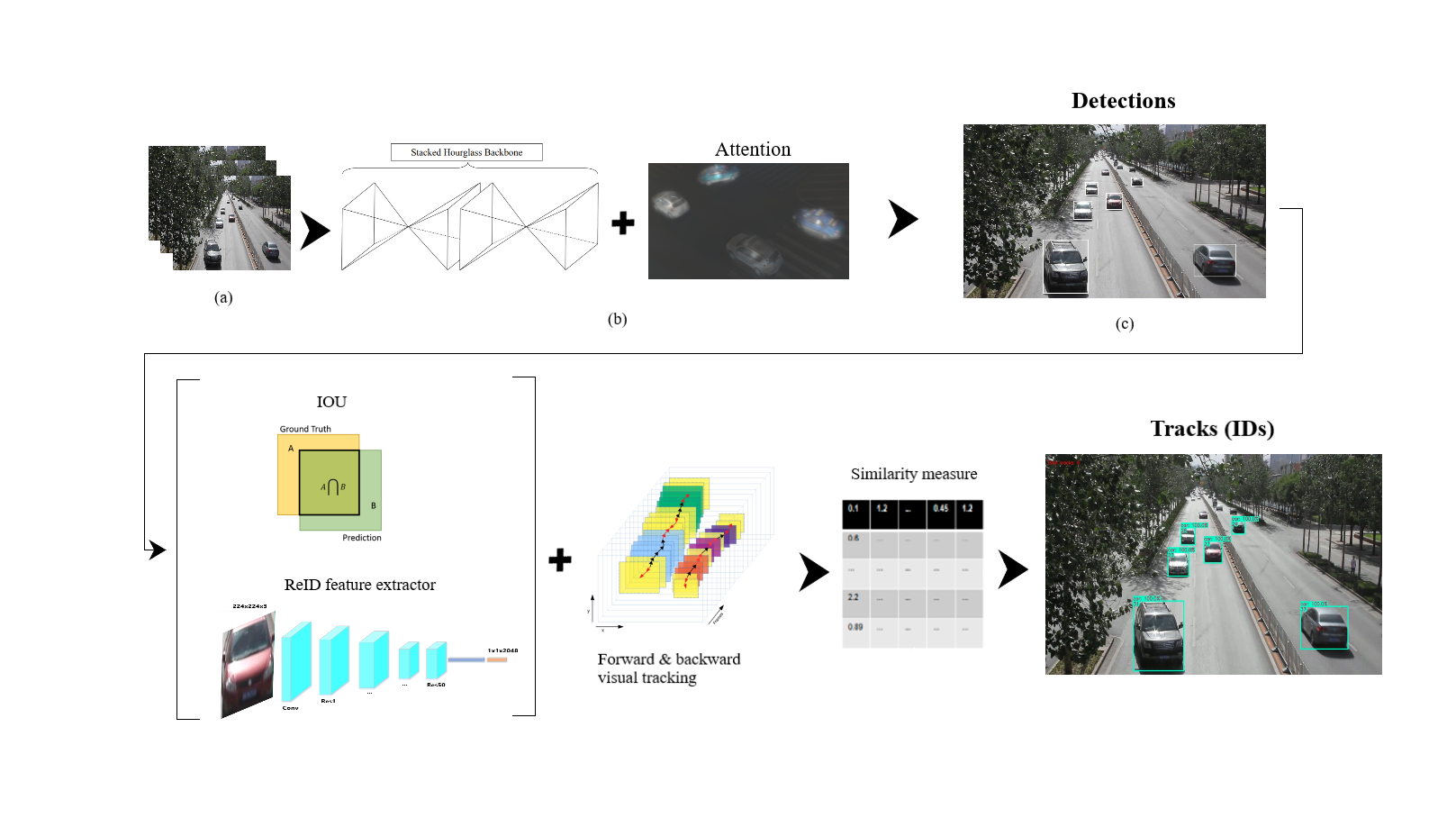}

\caption{Overall pipeline: (a) Our method takes as input RGB video frames. (b) Our detector of choice SpotNet \cite{perreault2020spotnet} generates candidate detections (bounding boxes, class labels and confidences) in each frame. (c) Detections are then passed to both the feature extractor and the association step. (d) Association costs are computed using the cosine similarity for each pair of ReID feature vectors and the IOU scores between bounding boxes. Visual object tracking is also employed forward and backward for unmatched tracks. (e) The final output is the established trajectories with their unique IDs.}
\label{fig:sct-workflow}

\end{figure}

We present the approaches we used for both the object detection and multiple object tracking components of our system in this section. An overview of our tracker is presented in Figure \ref{fig:sct-workflow}.

This work focuses on identifying vehicles in surveillance video sequences, then assigning unique IDs to each target and keeping track of them throughout the video stream. Given the good performance of such approaches in terms of MOT metrics and speed on numerous MOT challenges and benchmarks, as well as their modularity/adaptability to diverse applications, we chose a tracking-by-detection scheme to solve this problem. This means that our work is built around two key components: an object detection model for predicting vehicle locations and a multi-object tracker for linking detections across frames. 

\subsection*{Object detection}

We focused on selecting a model that can identify vehicles in urban scenes with a variety of constraints, such as occlusions and changing weather. As a result of these difficulties, object detection is a difficult task that necessitates ongoing study and the usage of relevant datasets tailored to the application. Our work confirms the idea that robust object detection, which enables effective object localization in difficult settings, is critical to the object tracking method overall performance.

For a good balance of speed and precision, we chose SpotNet \cite{perreault2020spotnet} as our vehicle detection model to recognize vehicles in video frames. Perreault et al. have proposed a model that utilises a CNN backbone, regression, and segmentation heads to create a simple and efficient detector. This model uses a dual Hourglass network composed of down-sampling and up-sampling layers to construct a rich feature map. The inclusion of a self-attention mechanism in this detector is its novelty, represented by multi-task learning, with object segmentation being learned as an additional task to guide the model. The segmentation branch was trained with semi-supervised segmentations obtained using PAWCS \cite{7045991} for background subtraction for fixed camera configurations, and Farneback optical flow \cite{10.1007/3-540-45103-X_50} for moving camera settings. On two object detection benchmarks with traffic scenarios, SpotNet outperformed the competition.

\subsection*{Base MOT method}

We build on the V-IOU tracker \cite{1547Bochinski2018} that follows the tracking-by-detection scheme in which the tracking method input is the object detection model bounding boxes, which supports our choice of a robust detector. This technique, like its predecessor \cite{1517Bochinski2017}, uses the IOU metric to assess the degree of overlap between previously detected bounding boxes and the predicted bounding boxes in the current frame to connect detections between frames. Then, it solves a linear assignment problem with a configurable minimum IOU threshold $\sigma_{IOU}$. When no association is detected, it reverts to visual object tracking to predict the next locations of a target within a time window defined by a parameter $TTL$, both forward and backward. We resort to MedianFlow \cite{10.1109/ICPR.2010.675} in our pipeline to temporarily track each unmatched target. The authors' findings show that V-IOU lowers fragmentation and ID switch rates compared to the base IOU tracker. However, this model has two flaws: it generates a moderately high number of false positives, which is likely due to the visual object tracker predictions, and it only uses spatial information from bounding boxes to make the association, which suggests a vulnerability in urban settings where vehicle appearances can change after being occluded.

\subsection*{Proposed ReID extension}

To address the problems above, we extend the matching process of the chosen V-IOU tracker with information characterizing the vehicle visual appearance to deal with long-term occlusions and large spatial displacements due to fast motion in order to develop a more efficient and robust MOT system. By minimizing fragmentation and ID switches, appearance cues have been shown to increase association outcomes.

Particularly, we chose to enhance the data association component of our MOT technique with vehicle ReID elements based on Miah et al.'s comprehensive analysis of different appearance features for MOT in urban scenarios \cite{miah2020empirical}. We utilize the feature extraction module from \cite{reid}, which has been trained on three vehicle ReID datasets. The CNN feature extractor is trained using Triplet loss and cross-entropy loss on a ResNet \cite{he2015deep} architecture with 50 layers. When training using classification annotations, cross-entropy was utilized so that the CNN could learn to minimize the classification loss amongst vehicle models, resulting in the acquisition of robust discriminative features. An adaptive feature learning approach presented in the same publication is also used to fine-tune the feature extraction model. From unlabeled video sequences, this method tries to extract positive and training examples to better adapt the visual domain to test sequences. 

We include the ReID feature in our tracker in the following way. Vehicle bounding boxes are output from the object detector and supplied to the feature extractor model, which generates feature vectors of size 2048 describing object appearances, as well as to the MOT method, which performs the data association step. Our new cost function is as follows:

\begin{equation}
    cost(d_i, t_j) = 1 - \alpha * IOU(d_i, t_j) - \beta * cosine(R_{d_i}, R_{t_j})
\end{equation}

for a detection $d_i$ and a track $t_j$, with $R_{d_i}, R_{t_j}$ respectively their 2048-dimensional ReID feature vectors, and $\alpha + \beta = 1$. The association cost of the bounding boxes is evaluated with the IOU, while the ReID feature vectors are compared with the cosine similarity. 

While visual object tracking can aid with short-term occlusions, ReID features are invariant to viewing angle and orientation change problems since they are derived from a CNN trained on numerous pictures of the same item with varied orientations and illumination conditions. The new association function will compute the spatial and appearance similarity between the anticipated bounding boxes and candidate detections when an item is occluded for a long duration. As a result, the object is more likely to be recovered after longer occlusions when utilizing the ReID model rather than simply employing IOU for matching. The anticipated locations and candidate detections would be far apart in cases where the vehicles travel at comparatively higher speeds, resulting in an IOU score near to zero. Because it helps minimize association costs, the ReID extension plays a key role in enhancing the chance of accurately matching predictions to detections in rapid motion scenarios. Backward predictions are also matched to terminated tracks using ReID vectors to generate more cohesive trajectories with fewer gaps. Finally, mismatched detections start new tracks, which are visually monitored until they are linked to fresh detections in the following $TTL$ frames, at which point they are terminated. Tracks shorter than $t_{min}$ are also eliminated. We include a custom cosine similarity function written in Python and accelerated using a package called Numba to keep the overall processing time of our system as low as feasible. This decreases the cosine affinity execution time to about $1/36th$ when compared to a conventional implementation.

\section{Results and Discussion}

\subsection*{Dataset and evaluation}

For all our fine-tuning and tests, we used the UA-DETRAC \cite{wen2020uadetrac} dataset and benchmark suite. It provides around 10 hours of video collected from real-world traffic settings at a frame rate of 25 FPS and a 960 x 540 pixels resolution. This results in more than 140,000 frames and a total of 1.21 million bounding boxes from 8250 vehicles. This dataset presents 4 vehicle classes (Car, Bus, Van, Other) and 4 weather conditions (Sunny, Cloudy, Rain, Night).

To examine the influence of the detection model on tracking, the evaluation procedure of UA-DETRAC processes detection and tracking data for each threshold in the range 0.0 to 1.0 with a step of 0.1. This implies that the detections with confidence scores less than or equal to the current threshold are chosen for each iteration, and the MOT technique is then assessed for this subset. The benchmark proposes a set of metrics called DETRAC-MOT which are PR-MOTA (MOT accuracy), PR-MOTP (MOT precision), PR-MT (mostly tracked), PR-ML (mostly lost), PR-IDS (ID switches), PR-FM (fragmentation), PR-FP (false positives) and PR-FN (false negatives). All of these metrics are based on the detector precision-recall curve.

We fine-tuned our proposed MOT method on the UA-DETRAC train split of 60 sequences. The best parameter values we obtained are: 

\begin{center}
    $\sigma_{IOU}: 0.6, \alpha: 0.7, \beta: 0.3, TTL: 15$ and $t_{min}: 3$
\end{center}

\subsection*{Results}

During our tests, we observed that SpotNet is rather safe with its confidence scores. It rarely outputs confidences higher than 90\%. So we multiplied the scores for the top 50 detections per frame by 1.25 and clipped them to 100\% to spread out the confidence score distribution between 0 and 1 to be compatible with the test procedure of UA-DETRAC. Finally, the results of our tracker shown in Table 2 confirm that our proposed method outperforms our baseline on the UA-DETRAC tracking benchmark, V-IOU with Mask R-CNN, particularly in terms of PR-MOTP with a 13.9\% increase. We also score a marginal improvement for PR-MOTA with a 0.5\% increase. Thus, we get state-of-the-art results of 31.2\% PR-MOTA and 50.9\% PR-MOTP. While the base V-IOU method still has the lead in terms of PR-MT, PR-IDS, PR-FM, we outperform it in terms of PR-ML, PR-FP and PR-FN. Altogether, our method is more consistent and better performing on most PR-MOT metrics. Our approach produces approximately three times less false positives than V-IOU with Mask R-CNN. This illustrates the considerable gain achieved by upgrading the object detector to SpotNet, a far more robust model capable of handling a variety of demanding circumstances. We also see a slight reduction in the number of false negatives.

On PR-ML, we see a noticeable improvement from 22.6 to 18.5. This measure quantifies the proportion of ground truth tracks that are tracked 20\% or less of the time. This supports the concept that integrating visual object tracking with ReID cues is a better method for improved trajectory maintenance, resulting in more consistent tracks overall. Nonetheless, we leave opportunity for additional studies with alternative values to fine-tune the weights of the spatial and appearance criteria $\alpha$ and $\beta$. 

\begin{table}[htb]
\centering
\caption{DETRAC-MOT results on the UA-DETRAC test set with top 50 SpotNet detections. \textbf{Bold}: best result in each metric. Results * taken from \cite{wen2020uadetrac}. Results \& taken from \cite{1547Bochinski2018}. Results \# taken from \cite{Kang2021}. The last line presents results of our method.}
\label{tab:viou+spotnet-MOT}
\begin{tabular}{|c|c|c|c|c|c|c|c|c|c|}
\hline
\textbf{Detector} & \textbf{Tracker} & \textbf{\begin{tabular}[c]{@{}c@{}}PR-\\ MOTA\end{tabular}} & \textbf{\begin{tabular}[c]{@{}c@{}}PR-\\ MOTP\end{tabular}} & \textbf{\begin{tabular}[c]{@{}c@{}}PR-\\ MT\end{tabular}} & \textbf{\begin{tabular}[c]{@{}c@{}}PR-\\ ML\end{tabular}} & \textbf{\begin{tabular}[c]{@{}c@{}}PR-\\ IDS\end{tabular}} & \textbf{\begin{tabular}[c]{@{}c@{}}PR-\\ FM\end{tabular}} & \textbf{\begin{tabular}[c]{@{}c@{}}PR-\\ FP\end{tabular}} & \textbf{\begin{tabular}[c]{@{}c@{}}PR-\\ FN\end{tabular}} \\ \hline
DPM               & GOG*              & 5.5                                                         & 28.2                                                        & 4.1                                                       & 27.7                                                      & 1873.9                                                     & 1988.5                                                    & 38957.6                                                   & 230126.6                                                  \\ \hline

EB               & KIOU*              & 21.1                                                         & 28.6                                                         & 21.9                                                        & \textbf{17.6}                                                      & 462.2                                                     & 712.1                                                   & 19046.9                                                   & \textbf{159178.3}                                                  \\ \hline
M-RCNN        & V-IOU\&            & 30.7                                                        & 37.0                                                        & \textbf{32.0 }                                                     & 22.6                                                      & \textbf{162.6}                                                      & \textbf{286.2}                                                     & 18046.2                                                   & 179191.2                                                  \\ \hline
SpotNet        & \begin{tabular}[c]{@{}c@{}}Kalman\\ +IOU\#\end{tabular}            & 30.6                                                        & 42.7                                                        & -                                                     & -                                                     & 3634                                                      & -                                                    & -                                                   & -                                                 \\ \hline \hline
      \textbf{SpotNet}            &          \begin{tabular}[c]{@{}c@{}}\textbf{V-IOU}\\\textbf{+ReID}\end{tabular}       &                   \textbf{31.2}                                      &          \textbf{50.9}                                                  &                                              28.1            &                                              18.5            &                                    252.6                    &                                                 329.1       &                                  \textbf{6036.4}                 &                                  170700.6                 \\ \hline
\end{tabular}
\end{table}

In Table \ref{tab:ablation}, we give a more detailed view about the impact of each step with an ablation study. When compared to utilizing SpotNet with the standard version of V-IOU, adding the ReID features substantially reduces the frequency of ID switches and fragmentation. Adding the ReID component results in the highest increase in PR-MOTA and PR-MOTP, while utilizing spotNet results in the biggest improvement in PR-FP. Through the design of our cost function, we interpret the gain in terms of PR-MOTP. Our cost is responsible for 30\% of the ReID feature similarity. The more we visually track an item using Medianflow \cite{10.1109/ICPR.2010.675} in circumstances with relatively long occlusions, the more inaccuracies are incorporated into the predicted boxes. As a result, IOU would assign low scores between the anticipated vehicle position and candidate detections. By lowering the association cost, the ReID scores will appropriately compensate for the IOU, allowing our technique to fill gaps in trajectories. It also helps in a similar way for detections that are spatially distant for smaller objects moving at fast speeds. This would result in extremely short tracks of length one, which would be rejected by the assessment process using the base V-IOU tracker. We can link these detections despite the displacement when ReID features are implemented.

\begin{table}[htb]
\centering
\caption{Ablation study on the UA-DETRAC test set. Bold: best result in each metric. Results \& taken from \cite{1547Bochinski2018}.}
\label{tab:ablation}
\begin{tabular}{|c|c|c|c|c|c|c|c|c|c|}
\hline
\textbf{Detector} & \textbf{Tracker} & \textbf{\begin{tabular}[c]{@{}c@{}}PR-\\ MOTA\end{tabular}} & \textbf{\begin{tabular}[c]{@{}c@{}}PR-\\ MOTP\end{tabular}} & \textbf{\begin{tabular}[c]{@{}c@{}}PR-\\ MT\end{tabular}} & \textbf{\begin{tabular}[c]{@{}c@{}}PR-\\ ML\end{tabular}} & \textbf{\begin{tabular}[c]{@{}c@{}}PR-\\ IDS\end{tabular}} & \textbf{\begin{tabular}[c]{@{}c@{}}PR-\\ FM\end{tabular}} & \textbf{\begin{tabular}[c]{@{}c@{}}PR-\\ FP\end{tabular}} & \textbf{\begin{tabular}[c]{@{}c@{}}PR-\\ FN\end{tabular}} \\ \hline
M-RCNN        & V-IOU\&            & 30.7                                                        & 37.0                                                        & \textbf{32.0 }                                                     & 22.6                                                      & \textbf{162.6}                                                      & \textbf{286.2}                                                     & 18046.2                                                   & 179191.2                                                  \\ \hline

SpotNet           & V-IOU           & 27.6                                                        & 39.8                                                  &                      23.6                      &                  20.9                                     &                 293.1                                     &         382.4                                           &          \textbf{4783.9}                                       &         194784.4                                 \\ \hline
      SpotNet            &         \begin{tabular}[c]{@{}c@{}}V-IOU\\ +ReID\end{tabular}         &                   \textbf{31.2}                                      &          \textbf{50.9}                                                  &                                              28.1            &                                              \textbf{18.5 }           &                                    252.6                    &                                                 329.1       &                                  6036.4                 &                                  \textbf{170700.6}                 \\ \hline
\end{tabular}
\end{table}

Finally, we present the processing speed values for each component of our system on the RTX 6000, with a detection threshold of 0.7. (yielding up to 20-25 candidates per frame). For the detection step, SpotNet can process up to 9 FPS, while our V-IOU with ReID and cosine extension can process up to \textbf{60 FPS}, resulting in a total end-to-end speed of up to 6 FPS.

\section{Conclusion}

Our work adopts the popular tracking-by-detection paradigm to propose an MOT method that combines spatial information with re-identification features to track vehicles in urban scenes in online use cases. Our data association function design shows solid performance on the UA-DETRAC benchmark for long-term occlusions and large shifts due to fast motion. This work enables further improvement in terms of speed as well as adding more information to further improvement tracking performance.


\end{document}